\UseRawInputEncoding
\RequirePackage{fix-cm}
\documentclass{svjour3}         
\smartqed  
\usepackage{graphicx}
\usepackage{epstopdf}
\usepackage{geometry}
\usepackage[numbers,sort&compress]{natbib}

\usepackage{bm}
\usepackage{amsmath,amssymb,amsfonts}
\usepackage{algorithm}
\usepackage[noend]{algpseudocode}
\usepackage{graphicx}
\usepackage{multirow}
\usepackage{etoolbox}
\usepackage{verbatim}
\usepackage{lineno}
\usepackage{setspace}
\usepackage{textcomp}
\usepackage[tight]{subfigure}
\usepackage{caption}
\usepackage{subfig}
\usepackage{epstopdf}
\usepackage[misc]{ifsym} 

\begin{document}
\title{{An advanced YOLOv3 method for small object detection} 
}


\author{Baokai Liu\textsuperscript{1}         \and
        Fengjie He\textsuperscript{2}        \and
        Shiqiang Du\textsuperscript{1,3}      \and
        Jiacheng Li\textsuperscript{1}         \and
        Wenjie Liu\textsuperscript{1}
}

\institute{%
	\begin{itemize}
		\item[\textsuperscript{\Letter}] {Shiqiang~Du } \\
		\email{shiqiangdu@hotmail.com}
		\at
		\item[\textsuperscript{1}] Key Laboratory of China's Ethnic Languages and Information Technology of Ministry of Education, Chinese National Information Technology Research Institute, Northwest Minzu University, Lanzhou, Gansu, 730030 China
		\at
		\item[\textsuperscript{2}] China Mobile Design Institute Co., Ltd. Shaanxi Branch, Xi'an, Shaanxi, 710000 China
      	\at
		\item[\textsuperscript{3}] College of Mathematics and Computer Science, Northwest Minzu University, Lanzhou, Gansu, 730030 China
	\end{itemize}
}


\date{Received: date / Accepted: date}

\maketitle
\begin{abstract}

Small object detection has important application value in the fields of autonomous driving and drone scene analysis. As one of the most advanced object detection algorithms, YOLOv3 suffers some challenges when detecting small objects, such as the problem of detection failure of small objects and occluded objects. To solve these problems, an improved YOLOv3 algorithm for small object detection is proposed. In the proposed method, the dilated convolutions mish (DCM) module is introduced into the backbone network of YOLOv3 to improve the feature expression ability by fusing the feature maps of different receptive fields. In the neck network of YOLOv3, the convolutional block attention module (CBAM) and multi-level fusion module are introduced to select the important information for small object detection in the shallow network, suppress the uncritical information, and use the fusion module to fuse the feature maps of different scales, so as to improve the detection accuracy of the algorithm. In addition, the Soft-NMS and Complete-IoU (CloU) strategies are applied to candidate frame screening, which improves the accuracy of the algorithm for the detection of occluded objects. The ablation experiment of the MS COCO2017 object detection task proves the effectiveness of several modules introduced in this paper for small object detection.
The experimental results on the MS COCO2017, VOC2007, and VOC2012 datasets show that the Average Precision (AP) of this method is 16.5\%, 8.71\%, and 9.68\% higher than that of YOLOv3, respectively.

\end{abstract}


\section{Introduction}
\label{sec:Introduction}
Object detection is an important branch of computer vision, which  purpose is to find the location of objects and complete the classification of objects. It has very important practical applications in many fields, such as monitoring safety, autonomous driving, traffic monitoring, scene analysis and robust vision, etc. In recent years, many deep learning based object detection methods have achieved state-of-the-art results, which can be roughly divided into one-stage and two-stage detectors.

The two-stage object detector first generates some object proposals on the image, which are then fed into the network for regression and classification. This type of method can achieve relatively high accuracy in object detection and detection tasks, but the inference speed will be relatively slow, and the purpose of real-time detection cannot be achieved.

In addition, the one-stage object detection method performs both regression and classification tasks directly on the predicted feature map, greatly improving the speed of detection. Among them, the YOLO network is the most representative type of one-stage object detection network \cite{2016You,chen2022object}. However, these methods have not achieved satisfactory results for the detection of small objects. This paper aims to solve the problem that the one-stage object detection network has unsatisfactory detection results for small objects \cite{chen2022small}.

The detection of small objects is very important for many robust vision tasks. In autonomous driving, detecting small or distant objects  in the high-resolution scene photos for the car is necessary in an effort to ensure self-driving cars safely, such as traffic signs or pedestrians. In industry, it is an effective method to find defects on the material surface  by using the method of small object detection.
Small objects have been smaller physical sizes in the real world.
It is generally defined in the MS COCO dataset as shown in Table \ref{tab:abc}. It can be seen that the small object has less pixel information. Due to the limited pixel features available for small objects, this will make detection very difficult.
For example, for an object with a size of 12*12 pixels, after 4 downsampling convolutions, the final generated feature map will not even have a feature point.
In addition, there are fewer images  that contain small objects in the dataset, which causes any detection model to focus more on medium and  large objects. These make the detection of small object relatively difficult.

In this study, a series of methods to improve the detection performance of small objects are proposed, such as data augmentation, context learning, and multi-level learning. These approaches inevitably have some limitations, such as the computational cost that data augmentation methods impose on the model. In recent years, Woo et al. \cite{woo2018cbam} present a lightweight attention module CBAM, which has been experimentally proved to have certain advantages in identifying target objects. Li et al. \cite{li2022outstanding} propose a YOLOv3 method of adaptive multi-level feature fusion to realize the detection of small objects in remote sensing images, which is not ideal for detecting small objects in general scenes because it does not consider the relationship between background and objects. Dong et al. \cite{dong2022lightweight} put forward a lightweight vehicle detection model based on YOLOv5, which proves that the CBAM module is of great significance for the selection of important features of vehicles.
Su et al. \cite{su2021pixel} develop a compact dilation convolution based module (CDCM) that effectively enriches multi-scale edge features in edge detection. Inspired by the above paper, a small object detection algorithm based on YOLOv3 is proposed.

The main contributions of our work can be  summarized as follows:

\begin{enumerate}
	\item  In the backbone extraction network, we use multi-level feature fusion module and CBAM to strengthen the fragile relationship between small objects and the background, and provide rich contextual semantic information for the prediction of small objects.

	\item  In this paper, a novel feature enhancement module DCM is proposed, which achieves the purpose of information enhancement in the predicted feature map by fusing the feature maps of different receptive field sizes.

	\item  An improved Soft-NMS strategy is applied to the model. By optimizing the matching and filtering rules for candidate boxes, we can increase the prediction accuracy for occluded objects while avoiding the issue of overlapping caused by directly reducing confidence scores.
	
\end{enumerate}

%
%
%

\begin{table}[t]
	\caption{Definition of objects size in MS COCO}
	\label{tab:abc}\centering
	\begin{tabular}{c|cc}   \hline
		\cline{1-3}
		   &Min scale area & Max scale area \\ \hline
		Small& $0 \times 0$ &$32 \times 32$    \\
		Medium &$32 \times 32$ &  $96 \times96$    \\
		Large &$96\times 96$ & $\infty\times \infty$ \\ \hline
	\end{tabular}
\end{table}

\begin{table}[t]
	\caption{Define  nouns and corresponding abbreviations}
	\label{tab:abc2}\centering
	\begin{tabular}{c|c}   \hline
		\cline{1-2}
		  noun & abbreviations \\ \hline
		 Convolutional Block Attention Module & CBAM \\
		 Dilated Convolutions Mish &  DCM    \\
            Complete-IoU &  CIoU \\
            Non-Maximum Suppression &NMS\\
             Average Precision &  AP\\
		mean Average Precision & mAP \\
 compact dilation convolution based module & CDCM \\
\hline
	\end{tabular}
\end{table}

\section{Related work}
\label{sec:related}
With the development of deep learning, people use deep learning methods to improve the detection accuracy of small objects. The research on deep learning methods is mainly carried out from the following four aspects.

\textbf{\subsection{Data Augmentation}}

Due to the uneven distribution of objects of various sizes in training samples, data augmentation is the most direct method to improve the performance of small object detection. Yu et al. \cite{yu2020scale} suggest a scale matching strategy, which reduces the loss of small object information by cropping different object sizes and reducing the difference between objects of different sizes. Kisantal et al. \cite{kisantal2019augmentation} present a method of replication enhancement, which increases the number of training samples of small objects by replicating small objects in the image. Accordingly, it improves the detection performance of small objects.
Chen et al. \cite{chen2019rrnet} propose an adaptive resampling strategy for data enhancement in RrNet by considering the context information of the objects, and achieve a better data augmentation result.

Data augmentation improves the detection performance of small objects by increasing the data samples of small objects. However, on the one hand, this method increases the amount of calculation. On the other hand, it introduces new noise by using inappropriate data augmentation strategies.

\textbf{\subsection{Multi-level Learning}}

Small objects have fewer pixel features. As the number of convolutional network layers increases, the feature and location information of small objects are gradually lost. The fusion of deep semantic information with shallow location information will improve the results of small object detection. Nayan et al. \cite{nayan2020real} demonstrate a real-time detection algorithm that extracts multi-level features of the network, which can improve the detection accuracy of small objects. Liu et al. \cite{liu2021hrdnet} propose a high-resolution detection network that retains the location information of small objects as much as possible to extract more semantic information. Deng  et al. \cite{deng2021extended} show an extended feature pyramid network for small object detection.

Multi-level learning effectively improves the detection performance of small objects by fusing shallow location information with deep semantic information.
However, this method is easily affected by noise in the process of fusing features.

\textbf{\subsection{Context Learning}}

In real life, objects are related to the background environment in which they are located. Through the study of this relationship, people model this relationship in deep learning to improve the performance of object detection.
Lim et al. \cite{lim2021small} propose  a method that combines context with multi-level features to improve the detection accuracy of small objects in the real world. Fu et al. \cite{fu2020intrinsic} put forward a new method based on the context, which is to solve the detection and missed detection of small objects by using more spatial information in the small objects.

This method makes full use of the relevant information between small objects and the background, and has a certain improvement in the detection of small objects. However, the lack of background information of the current object and the correlation between objects are not  considered.

\textbf{\subsection{Generative Adversarial Learning}}

The purpose of generative adversarial learning is to map low-resolution small objects to high-resolution objects. This method mainly improves the detection performance of small objects by increasing the features of small objects.
Bai et al. \cite{bai2018sod} develop a multi-task adversarial generative network that improves the detection performance of small objects by recovering clear objects from blurred small objects. Noh Bai et al. \cite{noh2019better}
propose a new super-resolution method that addresses the problem of mismatched receptive fields generating incorrect features.

Generative adversarial learning improves its detection performance by augmenting the features of small objects. However, the entire network training is more difficult and the generator cannot generate rich samples, so the performance improvement is limited.

YOLOv3 employs a feature pyramid network to predict objects of different sizes at different spatial resolutions.
For the detection of small objects, YOLOv3 shows a lower recognition rate on small objects because it does not consider the relationship between the object and the background and the problem of less information about small objects on the prediction feature map.
Inspired by multi-level learning and context learning models,
this paper uses CBAM in the backbone extraction network of the model to select the feature information of small objects and suppress non-important information or noise information in shallow features. In addition, our DCM module uses multi-level fusion of feature maps of different receptive fields to enrich feature information, and innovatively combines Soft-NMS and CIOU to improve the detection accuracy of occlusion small objects.

\section{Architecture of Network}
\label{sec:pre}

In this section, we will introduce the network architecture proposed in this paper in detail, and will introduce the network composition as a whole and give a detailed introduction to the different modules. Due to expand the following discussion more conveniently, some of the nouns used and the corresponding abbreviations are shown in Table \ref{tab:abc2}.

The method in this paper has made some improvements on the original YOLOv3 object detection network. The network is mainly composed of the backbone extraction network darknet-53, the feature enhancement module of the neck, and the prediction head. Since the original network did not consider the impact of network depth and neck feature fusion on small object detection, the detection accuracy of the original YOLOv3 network for small objects on the MS COCO2017 dataset was only 16.3\% \cite{lin2014microsoft}. Therefore, the method proposed in this paper mainly makes some improvements in the feature enhancement part of the original network and the backbone network \cite{li2022outstanding}. We added a feature enhancement module DCM to the backbone network of the network and a multi-level fusion module to the neck of the network to improve the detection accuracy of the network for small objects.
In addition, because small objects are easily occluded by large objects, the algorithm filters the candidate frames of small objects to a certain extent when screening candidate frames, which makes these small objects miss detection. To this end, we use the Soft-NMS and CIOU strategies on the one hand to reduce the penalty for small objects, and on the other hand to improve the detection accuracy of candidate boxes. Thus, the detection accuracy of occluded objects is improved.

\begin{figure*}[t]
	\centering
	\includegraphics[width=1\linewidth]{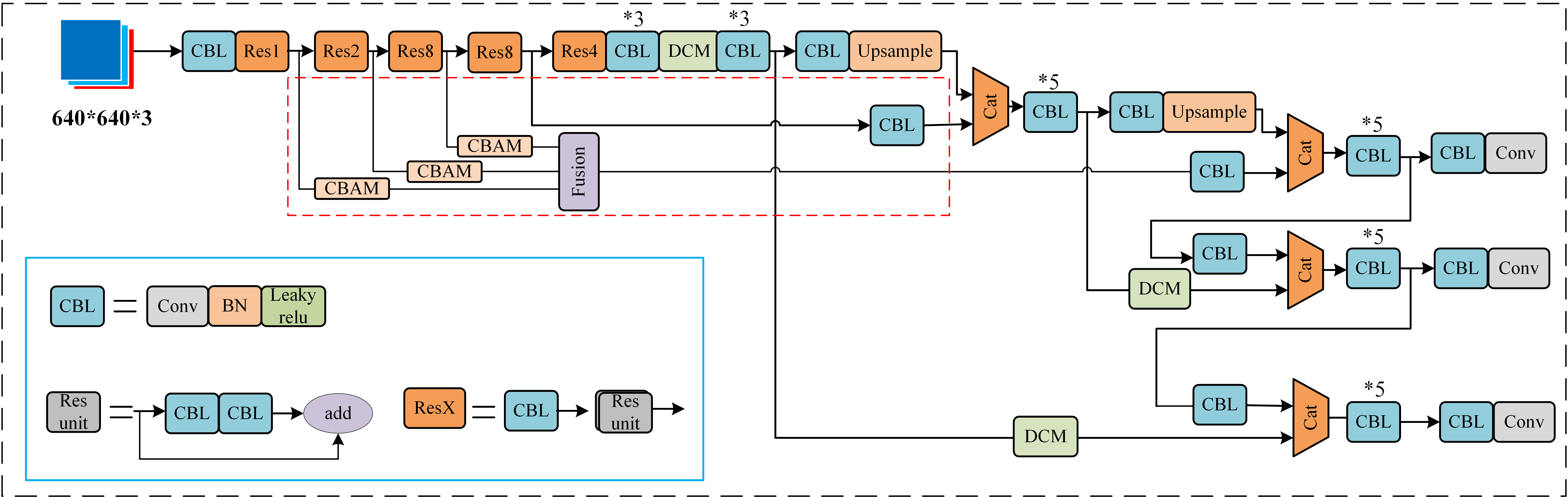}
	\hspace{0.1 cm}
	\caption{The proposed method based on the combination of feature fusion and feature enhancement can propagate the feature information of small objects in the shallow layer from the shallow layer of the network to the deep layer through skip connections. The CBAM module weights the shallow information through the attention module to reduce some shallow noise information, and the Fusion module mainly fuses the feature information of different scales extracted by the backbone network, and sends it to the small object detection head through the skip connection. The DCM module is mainly used in the backbone and neck of the network. The module in the backbone is mainly to enhance feature information, and the one in the neck is to expand the receptive field. }
	\label{fig:net}
\end{figure*}

Through the above analysis, we design the network structure shown in Figure \ref{fig:net}. The entire network structure is mainly composed of three parts: the backbone, the neck and the detection head. Among them, the backbone network uses multi-level fusion and attention mechanism to enhance the feature extraction ability of small objects. The neck structure strengthens the feature information extracted by the backbone network and expands the receptive field of the feature map by using the DCM module.
In this way, auxiliary information can be provided for the detection of small objects through the global feature information, which makes the detection of small objects more accurate.
For the detection head part, we use an improved detection head to predict the regression box and the classification task separately.

\textbf{\subsection{Attention Mechanism}}
The attention mechanism originates from the study of human vision. Humans give priority to important information and ignore unimportant information. In the convolutional neural network, the attention mechanism mainly uses the learned weight feature map to realize the screening of important information in the original feature map. We add an attention module to the multi-level feature fusion module at the neck of the network to reduce the interference of redundant information on small object prediction so as to improve the detection of small objects\cite{woo2018cbam}.

\begin{figure}[t]
	\centering
	\includegraphics[width=0.6\linewidth]{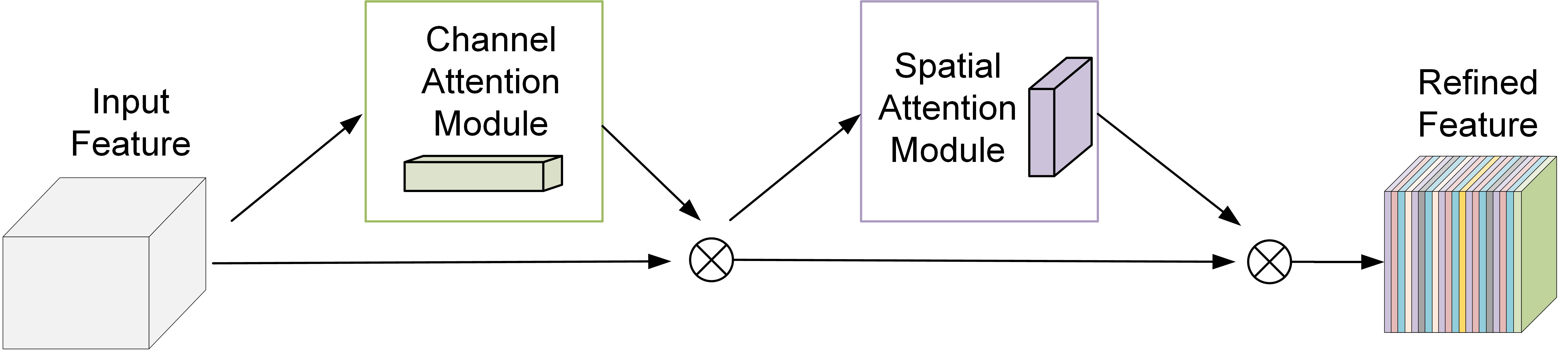}
	\caption{This module mainly includes channel attention and spatial attention. It is mainly composed of MLP layer network, which is used to weight the shallow semantic information and remove some noise information in the shallow feature map.}
	\label{fig:cbm}
\end{figure}

The structure of the module is shown in Figure \ref{fig:cbm}. The module combines spatial and channel attention. The output result of the convolutional layer will first pass a channel attention module to obtain different feature weights on the channel, and then pass through a spatial attention module, and finally  obtain the proportion result of the feature map.

The channel attention mechanism is to compress the feature map in the channel dimension to obtain different proportions of channel information and improve the feature representation ability. A channel description matrix is obtained by using global average pooling and global maximum pooling, and then sent to a fully connected network to obtain a weight matrix, and the final attention feature map is obtained by multiplying the weight matrix with the original feature map. The channel attention mechanism can be expressed as:
\begin{align}
M_c(F) & \nonumber =\sigma(MLP(AvgPool(F))+MLP(MaxPool(F))) \\
&=\sigma(W_1(W_0(F_{avg}^c))+W_1(W_0(F_{max}^c)))
\end{align}
where $\sigma$ denotes the sigmoid function, $W_0 \in \mathbb{R}^{C/r \times C} $, and $W_1 \in \mathbb{R}^{C\times C/r}$. Note
that the MLP weights, $W_0$ and $W_1$ are shared for both inputs and the ReLU activation function is followed by $W_0$.

The spatial attention weights different regions in the image to find the most important parts in the network for processing.
The specific operation process is as follows:
for a feature map with an input size of~$H\times W\times C$~. First, the max pooling and average pooling operations are performed on the spatial dimension to obtain two feature maps of size~$H\times W\times 1$. Then, the two feature maps are spliced, and the weight coefficients are obtained through a 3*3 convolutional layer. Finally, the weight coefficient is multiplied with the input feature map to obtain a new weighted feature map.

\begin{align}
M_s(F) & \nonumber =\sigma(f^{3\times 3}([AvgPool(F);MaxPool(F)])) \\
&=\sigma(f^{3\times3}([F_{avg}]^s;F_{max}^s))
\end{align}
where $\sigma$ denotes the sigmoid function and $f^{3 \times 3}$ represents a convolution operation with the filter size of $3\times3$.

\textbf{\subsection{Feature Enhancement Module}}

Generally, after the downsampling operation of the first few convolutional layers in the backbone extraction network, the resolution of the feature map has become very low. If the downsampling operation is continued, a lot of context information will be lost. Inspired by dilated convolution and U-net network, we propose a new feature enhancement module DCM \cite{misra2019mish}.
The downsampling operation is no longer used in this module, and this module is applied to the deep network of the backbone extraction network and the neck feature enhancement module respectively. The useful information in the prediction feature map is enriched by fusing and enhancing multiple features in the feature map, and the enhanced feature map and the prediction feature map are effectively fused to provide more information for the small object prediction feature map, thus It effectively improves the prediction accuracy of small objects.

\begin{figure}[t]
	\centering
	\includegraphics[width=0.6\linewidth]{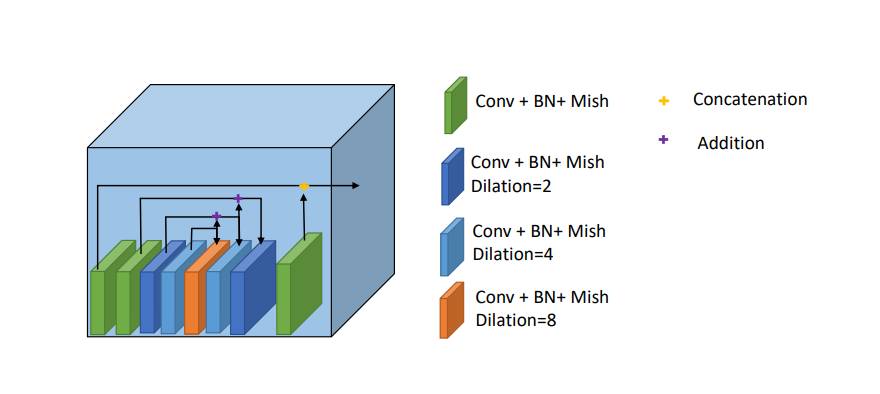}
	\caption{This is our proposed feature enhancement module, which is mainly composed of convolutional layers with different dilation rates, and enhances the feature information of the input feature map by adopting cross-layer fusion. Finally, the input feature information is fused with the module output information in a way similar to residual connection.}
	\label{fig:dcm}
\end{figure}

The main structure of the module is shown in Figure \ref{fig:dcm}.
It is composed of a set of 3*3 convolutions with different expansion coefficients, and uses symmetrical structures with expansion coefficients of 2, 4, 8, 4, and 2 to obtain feature information under different receptive fields. The feature maps under these different receptive fields are used to perform corresponding feature fusion using a network structure similar to U-net. For small objects, this structure can effectively provide more context information for the prediction of small objects, thereby improving the detection accuracy of small objects.

\textbf{\subsection{Multi-level feature fusion}}
For the entire network, low-level features contain more information about small objects, such as spatial and location information, but they  lack  more semantic information and contain more other noises. However, high-level features contain more abstract semantic information, which is beneficial for noise suppression. Therefore, we fuse the features of the two levels to obtain richer feature information.
In order to prevent the noise information in the low-level features from entering the final prediction feature map, we use the CBAM module to filter the noise information. We use a multi-level fusion module to fuse and complement the three low-level feature maps to preserve the spatial information in small objects. Specifically, the different features of the first three residual blocks of the backbone network are concatenated in the channel dimension.

\begin{figure}[t]
	\centering
	\includegraphics[width=0.6\linewidth]{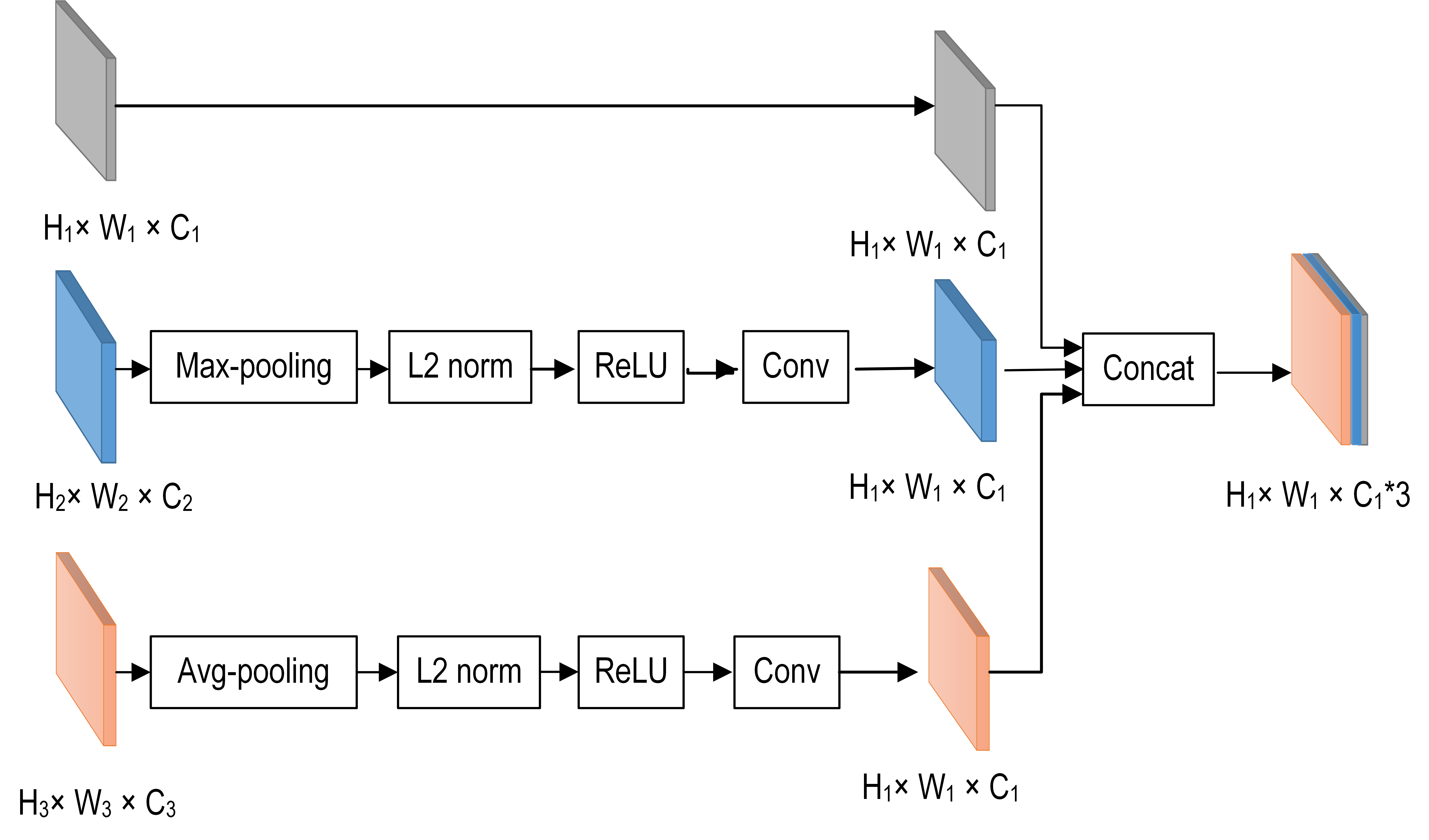}
	\caption{The fusion module mainly consists of three paths, where the feature maps of the lower two paths are downsampled using average pooling and max pooling, and then L2 regularization is used to balance different samples. Finally, the module uses 1*1 convolution to convert the number of channels to the specified size, and splices in the channel dimension to obtain the final output result.}
	\label{fig:fusion}
\end{figure}

The backbone network outputs feature maps with different scales, such as 80*80, 40*40, and 20*20. The 80*80 feature map is used to predict small objects. In the process of forwarding convolution of the network,
the feature map will gradually become smaller, and the information of the shallow features cannot be well retained in the deep feature map.
As a result, the feature information of the upper layer cannot be effectively transmitted to the next layer. We add a new structure with the aim of solving this problem. This structure is shown in Figure \ref{fig:fusion}. The structure contains a max pooling layer, an average pooling layer and a convolutional layer. We feed the three different outputs into the CBAM module to get the weighted feature map.
Then, we downsample the output feature map of the first residual block 1 times using max pooling and average pooling to downsample the output feature map of the second residual block 2 times. Then, we feed the three different outputs into the CBAM module to get the weighted feature map. And finally, we concatenate the three weighted feature maps in the channel dimension. In this feature fusion module, we use $L2$ regularization instead of batchnorm is mainly to balance the differences between samples \cite{he2015spatial}. Through this structure connection, the information of the upper layer of the network can be more transmitted to the network structure of the next layer \cite{liu2018path,wang2020cspnet}. The above process can be described as
\begin{align}
  \bm{\hat{f_i}}=cbam(\bm{f_i}) ~~~~~~~ i=1,2,3
\end{align}
\begin{align}
 \bm{f_{i}^t} = conv_1(\delta(downsample_j(\bm{\hat{f_i}})))~~~~i,j=1,2
\end{align}
where $\bm{\hat{f_i}}$ denotes the weighted low-level features, $\delta$ denotes the ReLU activation function, i is the residual block index, j is downsampling times, $conv_1$ is a 1*1 convolution layer, output dimension of 256 channels.

Finally, these three level features are concatenated in channel and then get final features by using a 3*3 convolution layer:
\begin{align}
\bm{f_{a}^t}=conv_2(concat(\bm{f_{1}^t},\bm{f_{2}^t},\bm{\hat{f_3}}))
\end{align}

\begin{figure}[t]
	\centering
	\includegraphics[width=0.6\linewidth]{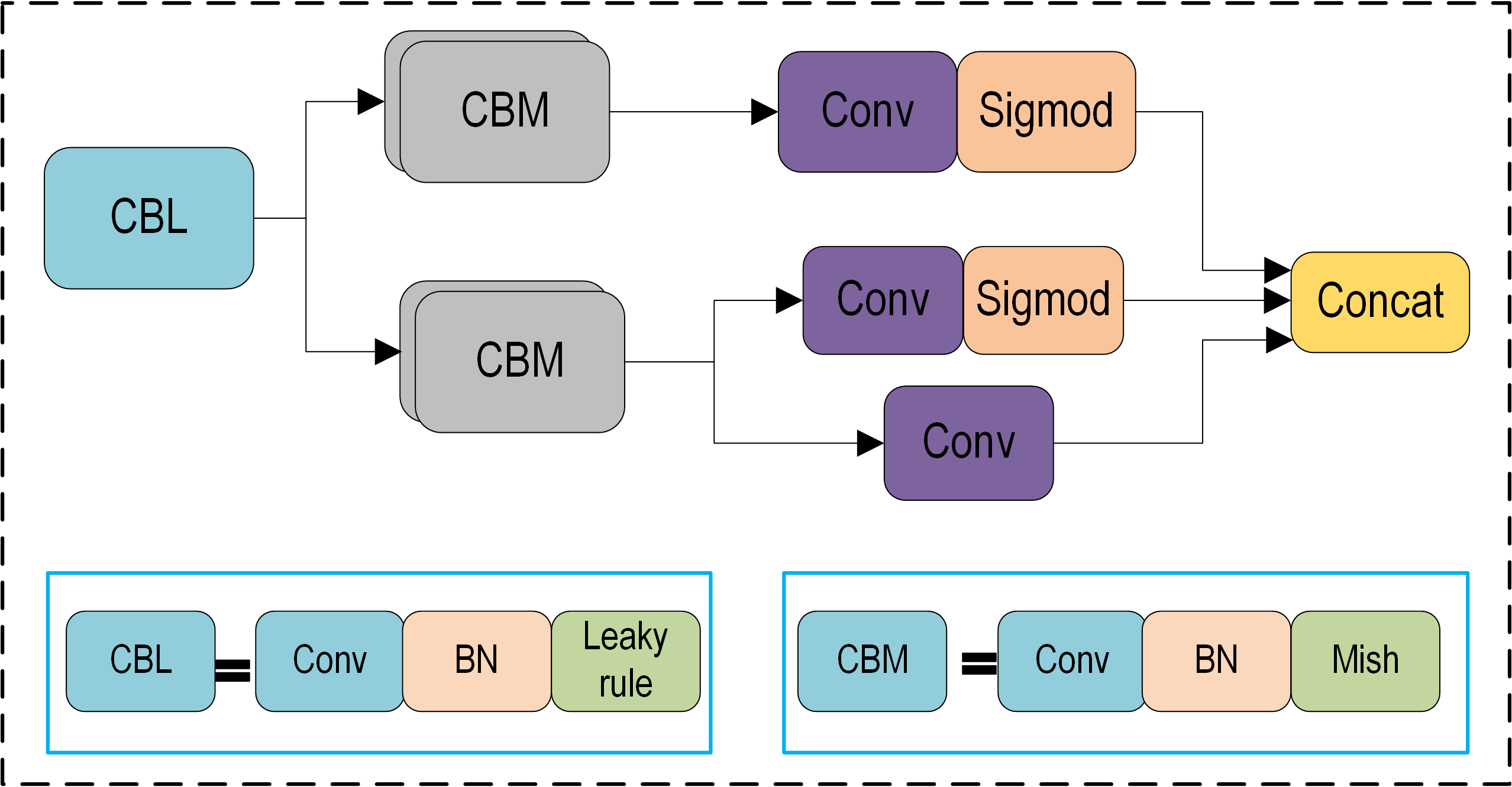}
	\caption{This module is mainly composed of two branches, the upper one is the classification branch, and the lower one is the regression branch. The blue CBL module is a 1*1 convolution, which mainly reduces the number of channels of the feature, and the gray CBM module is composed of two 3*3 convolutions, which are used to complete classification and regression tasks. The purple module is a 1*1 convolution that mainly reduces the number of channels to a specific number, while the yellow module splices the results of classification and regression according to the channel dimension.}
	\label{fig:head}
\end{figure}

\textbf{\subsection{Detection Head}}
The detection head of YOLOv3 is very simple. It consists of a 1*1 convolutional layers and a 3*3 convolutional layers to  obtain the final predictions. In the network, the regression and classification tasks will affect each other, and predicting the two tasks together will have a greater impact on the final result. We replace the YOLOv3 detect head
with decoupled head as in Figure \ref{fig:head}. it contains a 1*1 convolutional layer and two 3*3 convolutional layers. The 1*1 convolutional layer is mainly to reduce the channel dimension and the two 3*3 convolutional layers are used for regression and classification prediction tasks respectively. The first branch is mainly used to predict the category of objects. The other branch is mainly used to predict the size and position of the object regression box.
We replace the leaky relu function in the 3*3 convolutional layer with the Mish activation function, which can further improve the accuracy and generalization of the network detection head \cite{ge2021yolox}.

\textbf{\subsection{Soft-NMS}}
Non-maximum suppression is a very important part of the object detection network. First, it will sort according to the intersection score of the prediction box and the labeled box, and select the detection box with the highest score. Then, it performs the IOU calculation with other prediction boxes, and if it is greater than the set threshold, then discard this prediction box. Finally, from the remaining prediction boxes, it finds the one with the largest score, and so on. However, the NMS will have problems with the detection of dense objects, because the NMS will force the scores of adjacent detection boxes to zero. In this case, multiple overlapping objects will fail to detect, which will reduce the average detection accuracy of the network. Therefore, we can reduce the score of the adjacent detection boxes instead of completely removing it. Although the score is reduced, the adjacent detection boxes is still in the sequence of object detection. This is Soft-NMS \cite{bodla2017soft}. Algorithm \ref{alg:rtf2} shows the detailed steps.

\begin{algorithm}[htb]
	\caption{Algorithm of soft-NMS} \label{alg:rtf2}
	\begin{algorithmic}[1]
\State \textbf {Input:}$B={b_1,...,b_N},S={s_1,...,s_N},N_t$~
\State~~~~~~~~~~~$\bm{B}$ is the list of initial detection boxes
\State~~~~~~~~~~~$\bm{S}$ contains corresponding detection scores
\State~~~~~~~~~~~$N_t$ is the NMS threshold\\
\vspace{0.1cm} \textbf {begin}\\
\vspace{0.1cm}~~~~$D\leftarrow\{\}$\\
\vspace{0.1cm}~~~~\textbf {while}~~B$\neq$ empty~~\textbf{do} \\
\vspace{0.1cm}~~~~~~~~~~$m\leftarrow$ argmax \textbf {S}\\
\vspace{0.1cm}~~~~~~~~~~$M\leftarrow b_m$\\
\vspace{0.1cm}~~~~~~~~~~$D\leftarrow D\cup M;_B \leftarrow B-M$\\
\vspace{0.1cm}~~~~~~~~~~\textbf {for}~ $b_i$~in~ B ~\textbf{do}\\
\vspace{0.1cm}~~~~~~~~~~~~~~~~$s_i\leftarrow s_i f(ciou(M,b_i))$\\
\vspace{0.1cm}~~~~~~~~~~\textbf{end}\\
\vspace{0.1cm}~~~~\textbf{end}\\
\vspace{0.1cm}~~~~\textbf{return} D, S\\
\vspace{0.1cm}\textbf{end}\\
	\end{algorithmic}
\end{algorithm}
Most of the small objects in the dataset cannot be adequately detected due to being occluded by large objects. we choose to use the CIOU to replace the IOU in the original algorithm \cite{zheng2020distance}.
Because the CIOU considers the distance between the object and the anchor box, the repetition rate, and the aspect ratio, which can make the prediction frame of each object more accurate and avoid some unnecessary coincidence of prediction frames occurs. The formula is as follows:
\begin{align}
	\label{eq:x1}
	L_{CIOU}=1-IOU+\frac{\rho^2(\bm{b},\bm{b}^{gt})}{c^2}+\alpha \nu
\end{align}
\begin{align}
	\label{eq:x2}
	\nu =\frac{4}{\pi^2}(\arctan \frac{w^{gt}}{h^{gt}}-\arctan \frac{w}{h})^2
\end{align}
where $\bm{b}$ denotes  bounding box, $\bm{b}^{gt}$ denotes
ground truth of bounding box, c denotes the diagonal distance between $\bm{b}$ and $\bm{b}^{gt}$ smallest bounding rectangle, $\rho(.)$ euclidean metric. w, h denote length and width of the bounding box. $w^{gt}$, $h^{gt}$ denote length and width of the bounding box. $\alpha$ denotes weights.

For the penalty function, it is best to choose a continuous one, otherwise it will cause a sudden change in the detection order. Continuous penalty functions should have no penalty when there is no overlap, and should have a high penalty when there is high overlap. In addition, when the degree of overlap is low, the penalty should be gradually increased, that is, M should not affect the score of the prediction box with a low degree of overlap. However, when the coincidence of $B_i$ and M is close to 1, $B_i$  will add a significant penalty. Based on this consideration, we choose the Gaussian penalty function, the function expression is as shown in formula (\ref{eq:x0}), it uses this function to update each iteration process, and update the scores of all remaining detection frames, and finally found through experiments, the $N_t$ value Set to 0.9, g=1.1, the best experimental effect is obtained.
\begin{align}
\label{eq:x0}
s_i=s_ie^{-\frac{ciou(M,B_i)^2}{\delta}},\forall B_i \not \in D
\end{align}

\textbf{\subsection{DataSet}}
\textbf {Training dataset:} The MS COCO is a dataset provided by the Microsoft team for a variety of deep learning tasks, including object detection, key point detection, instance detection, and panoramic segmentation. Among them, the COCO2017 dataset includes 118287 training sets, 5000 validation sets, and 40670 test sets.


\textbf {Evaluation datasets:} The PASCAL VOC dataset and the MS COCO dataset are selected to test the model. The VOC2007 dataset includes 9963 pictures, including 24640 objects marked. The VOC2012 dataset contains 11540 images, and 27450 objects are labeled.
We randomly selected 4952 and 5000 pictures in the VOC dataset and the COCO dataset as the test set for the experiment, respectively.


\textbf{\subsection{Evaluation Metrics}}

In order to quantitatively evaluate the performance of the model proposed in this paper, we use the average accuracy of the evaluation index commonly used in object detection to measure the effect of the model. For the purpose of more comprehensively evaluating the performance of different models, we choose different IOU, and then the mean value of AP under these IOU.\\
(1) \textbf{IOU} is used to calculate the ratio of the intersection and union of two bounding boxes. In general, the higher the IOU value, the more accurate the object detection. The formula is defined as follows:
\begin{equation}
	IOU=\frac{intersection}{Union}
\end{equation}
(2)\textbf{AP} refers to the average accuracy rate, which averages the accuracy at different recall rate points. The larger the AP value, the higher the average accuracy of the model.
In this paper, we set the threshold of IOU to 0.5, and calculate the final AP value by taking 10 points on the PR curve. The formula is expressed as follows:
\begin{equation}
	AP=\frac{1}{11}\sum_{r \in {0,0.1,...,1}}P_{interp}(r)
\end{equation}
(3)\textbf{mAP}. The dataset contains multiple object categories, and an AP value is calculated for each category. In order to comprehensively compare the performance of the algorithm, the result obtained by averaging the AP values of all categories is called mAP (mean Average Precision).

\begin{equation}
	mAP=\frac{\sum^N_{i=1} AP_i}{N}
\end{equation}
\textbf{\subsection{Implementation Details}}

In order to prove the effectiveness of the method proposed in this paper, we implemented, trained and verified the model on the pytorch platform. Among them, we selected COCO-train 2017 as the training data set, and resized each picture into a scale of 640*640. In addition, some hyper parameters in network training (learning rate lr=1e-5, batch\_size=16, weight\_decay=5e-4) are set. For the purpose of  fast training of the network, this paper conducts two-stage training on two 2080Ti GPUs. In the first stage, during the first 50 epochs, the entire model backbone is trained frozen. In the second stage, in the following 150 epochs, the backbone network of the model is thawed and participates in the training of the entire network. Finally, the network tends to converge after about 200 hours of training.

\textbf{\subsection{Experimental results}}
To compare the performance of the algorithm in this paper and several other commonly used object detection algorithms in small object detection tasks, we selected MS COCO2017 and VOC datasets and used the COCO API toolbox for performance testing.
The evaluation results of all algorithms are shown in Table \ref{tab:coco}, Table \ref{tab:voc2007} and Table \ref{tab:voc2012}.
As can be seen from  Table \ref{tab:coco}, the algorithm in this paper has an AP value of at least 5.4\% higher than other algorithms in small object detection. For the detection of large objects, the AP value of the algorithm is also increased by about 3.5\%. For the VOC dataset, the algorithm in this paper improves by about 2\%.



\begin{figure*}[htbp]
	\centering
	\subfigure[YOLOv3.]{
	\begin{minipage}[t]{0.25\linewidth}
		\centering
		\includegraphics[width=1.5 in]{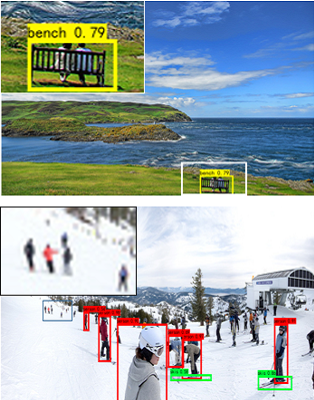}
	\end{minipage}%
	}%
	\subfigure[YOLOv4.]{
	\begin{minipage}[t]{0.25\linewidth}
		\centering
		\includegraphics[width=1.5 in]{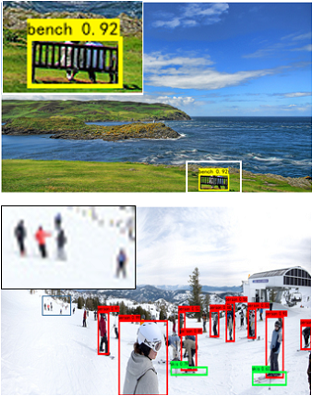}
	\end{minipage}%
	}%
	\subfigure[YOLOv5.]{
	\begin{minipage}[t]{0.25\linewidth}
		\centering
		\includegraphics[width=1.5 in]{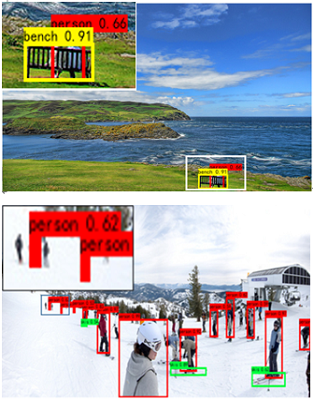}
	\end{minipage}%
	}%
	\subfigure[Ours.]{
	\begin{minipage}[t]{0.25\linewidth}
		\centering
		\includegraphics[width=1.5 in]{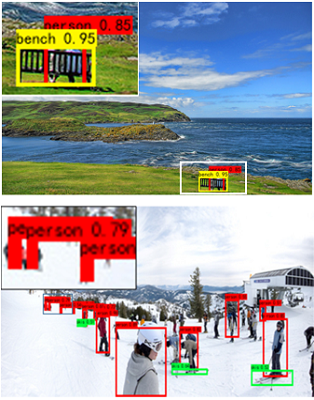}
	\end{minipage}%
	}%
	\centering
	\caption{Our proposed method can better capture the location of small objects in the picture compared with several other methods. The YOLOv3 and YOLOv4 object detectors have been unable to detect relatively small people, while the YOLOv5 object detector has a better performance, but there is some deviation in the location of small people.}
	\label{fig:result}
\end{figure*}

In order to more intuitively demonstrate the accuracy of the algorithm in this paper for small object detection, we visualized the detection results of different algorithms, and the results are shown in Figure \ref{fig:result}.
Among them, the YOLOv3 and YOLOv4 algorithms miss detection of distant characters in the picture, and cannot accurately detect occluded or relatively small objects. The main reason may be due to the lack of sufficient contextual semantic information for small objects in the network, which affects the detection of small objects. For the YOLOv5 algorithm, some small objects in the picture can be identified, but there is a problem of detection failure for some occluded objects. This is mainly because the YOLOv5 algorithm uses the method of small object data enhancement during network training, which improves the detection accuracy of the network for small objects. However, since the network does not consider the context semantics of small objects, there will still be missed detection problems for some occluded objects. However, the method in this paper can distinguish and detect small objects and occluded objects in the picture very well.

\begin{figure*}[htbp]
	\centering
	\subfigure[Image]{
	\begin{minipage}[t]{0.33\linewidth}
		\centering
		\includegraphics[width=2.1 in]{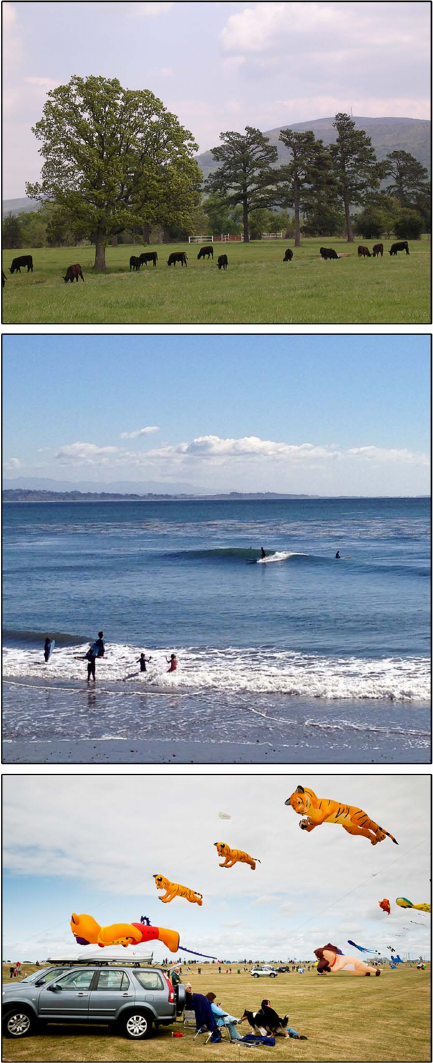}
	\end{minipage}%
	}%
	\subfigure[YOLOv3.]{
	\begin{minipage}[t]{0.33\linewidth}
		\centering
		\includegraphics[width=2.1 in]{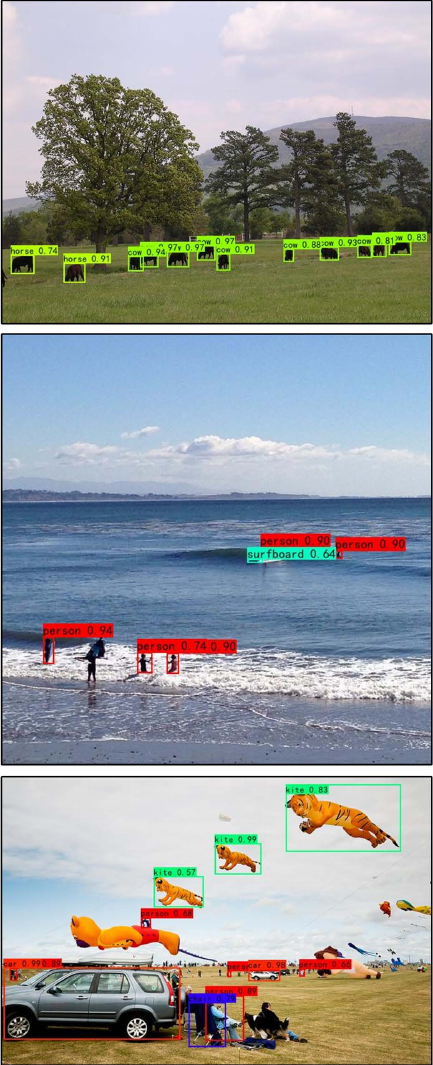}
	\end{minipage}%
	}%
	\subfigure[Ours.]{
	\begin{minipage}[t]{0.33\linewidth}
		\centering
		\includegraphics[width=2.1 in]{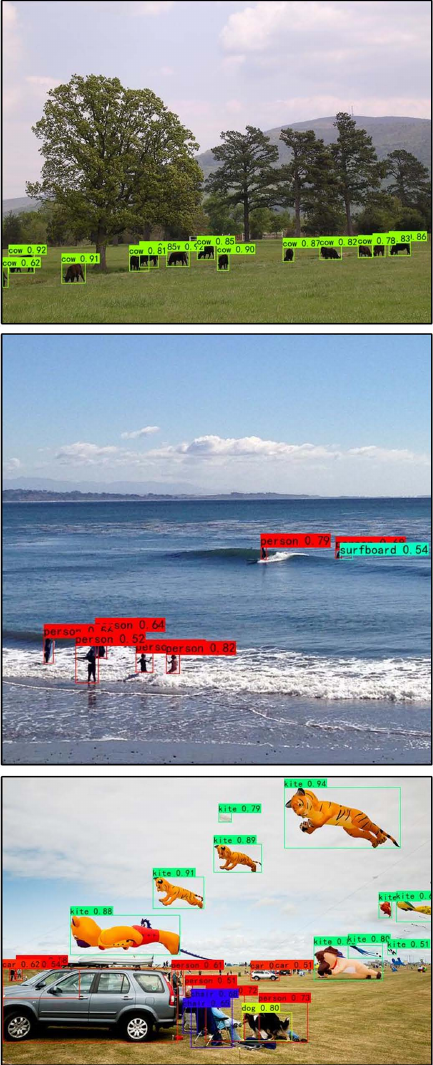}
	\end{minipage}%
	}%
	\centering
	\caption{Comparison of the experimental results of the YOLOv3 algorithm with the improved YOLOv3 algorithm}
	\label{fig:result2}
\end{figure*}

We can see from the experimental results in the first and second rows in Figure \ref{fig:result2} that the YOLOv3 detection algorithm has the problem of detection errors and missed detection of small objects occluded due to occlusion between objects. This is mainly because the YOLOv3 algorithm uses non-maximum suppression, and there are problems of inaccurate matching and misidentification of candidate frames for adjacent objects when screening candidate frames.The algorithm in this paper improves the positioning of candidate frames and retains redundant candidate frames by combining CIoU and Soft-NMS to achieve the detection of occluded objects.In addition, the third row of the figure can be seen that the YOLOv3 algorithm has the problem of detection failure for small objects with a single background, which is mainly due to the fact that the algorithm ignores the relationship between the background and the object and the reason why the predicted feature map information is less, and the algorithm in this paper uses CBAM to capture the relationship between the object and the background and the object and strengthens the information of the predicted feature map by using DCM and multi-level fusion, so as to achieve the detection of these small objects.

\begin{table*}[t]
	\caption{Accuracy($\%$) of different object detectors on the MS COCO2017 dataset.}
	\label{tab:coco}\centering
	\begin{tabular}{ccccccccc}  \hline
		\cline{1-7}
		Method        &AP$_{50:95}$     &AP$_{50}$                &AP$_{75}$       &AP$_{S}$               &AP$_{M}$          &AP$_{L}$  &parameter  &Time$_{GPU}$(sec.)   \\ \hline
		YOLOv3 \cite{redmon2018yolov3}  &$38.5$  &$69.3$        &$39.7$         &$16.3$       &$37.9$     &$53.2$     &235M   &6.47\\
		RetinaNet	&$40.8$  &$61.1$ &$44.1$   &$24.1$ &$44.2$
		&$51.2$   &145M &0.083 \\
        STDN \cite{zhou2018scale}	&$31.80$  &$51.10$ &$33.60$   &$14.40$ &$36.10$
		&$43.40$   &$-$ &$-$\\
        SSD	\cite{2016SSD}&$31.20$  &$50.40$ &$33.30$   &$10.20$ &$34.50$
		&$49.80$   &$-$ &$-$\\
        RefineDet \cite{zhang2018single}	&$33.00$  &$54.50$ &$35.50$   &$16.30$ &$36.30$
		&$44.30$   &$-$ &$-$\\
         LocalNet\cite{yan2021detection}	&$34.40$  &$53.20$ &$36.90$   &$20.30$ &$42.30$
		&$47.10$   &$-$ &$-$\\

		Faster R-CNN \cite{girshick2015fast} &$36.8$	 &$57.7$  &$39.2$ &$16.2$ &$39.8$
		&$52.1$ &160M &0.82\\
		YOLOv4 \cite{bochkovskiy2020yolov4}  &$43.5$   &$65.7$   &$47.3$  &$26.7$  &$46.7$  &$53.3$  &245M &0.317\\
		YOLOv5-m &$43.6$   &$62.3$   &$48.1$  &$27.4$  &$\bm{49.2}$
		&$55.7$ &41.1M  &$\bm{0.024}$ \\
		YOLOv3 ours   &$ \bm{47.1}$    &$\bm{76.3}$   &$\bm{48.5}$  &$\bm{32.8}$   &$46.2$
		&$\bm{59.2}$  &257M &1.73\\		
		
		\hline
	\end{tabular}
\end{table*}

\begin{table*}[t]
	\caption{Accuracy($\%$) of different object detectors on the VOC2007 dataset.}
	\label{tab:voc2007}\centering
	\begin{tabular}{cccccccccccc}  \hline
		\cline{1-12}
		Method     &bicycle   &bird   &boat   & bottle   &bus    & car  &cat
		& chair  &cow  & dog   & mAP \\ \hline
		YOLOv3 \cite{redmon2018yolov3}  &$86.97$  &$83.41$   &$53.59$    &$73.41$  &$91.64$   &$83.19$  & $89.93$
		  & $69.64$   & $84.36$    & $84.13$     &$81.31$    \\
		SSD \cite{2016SSD}	 &$81.23$  &$83.41$   &$73.27$    &$52.14$  &$82.14$   &$83.63$  & $87.32$
		& $53.68$   & $77.74$    & $83.51$     &$74.35$       \\
		Faster R-CNN \cite{girshick2015fast} &$-$	 &$-$  &$-$ &$-$ &$-$ &$-$ &$-$ &$-$ &$-$ &$-$
		 	&$71.51$\\
        MONET \cite{gong2019using} &$-$	 &$-$  &$-$ &$-$ &$-$ &$-$ &$-$ &$-$ &$-$ &$-$
			&$83.00$\\
        BlitzNet \cite{dvornik2017blitznet} &$-$	 &$-$  &$-$ &$-$ &$-$ &$-$ &$-$ &$-$ &$-$ &$-$
		 	&$80.70$\\
        RefineDet \cite{zhang2018single} &$-$	 &$-$  &$-$ &$-$ &$-$ &$-$ &$-$ &$-$ &$-$ &$-$
		 	&$81.80$\\
        LocalNet \cite{yan2021detection} &$-$	 &$-$  &$-$ &$-$ &$-$ &$-$ &$-$ &$-$ &$-$ &$-$
		 	&$82.90$\\

		YOLOv4 \cite{bochkovskiy2020yolov4}  &$85.60$  &$81.43$  &$54.75$
		&$74.33$   &$93.55$  &$81.59$ &$88.40$
		&$69.70$  &$88.78$ & $85.54$  & $82.44$\\
		YOLOv5-m   &$92.10$   &$91.02$   &$77.42$  &$80.55$  &$\bm{96.49}$  &$86.73$
		& $92.83$ &$75.64$   &$\bm{95.74}$   &$90.91$    &$88.16$\\

		YOLOv3 ours &$\bm{92.72}$   &$\bm{91.22}$   &$\bm{83.76}$  &$\bm{84.74}$  &$94.68$ &$\bm{96.42}$ &$\bm{93.86}$ &$\bm{78.23}$ &$93.56$ &$\bm{92.75}$
&$\bm{90.02}$   \\
		\hline
	\end{tabular}
\end{table*}

\begin{table*}[t]
	\caption{Accuracy($\%$) of different object detectors on the VOC2012 dataset.}
	\label{tab:voc2012}\centering
	\begin{tabular}{cccccccccccc}  \hline
		\cline{1-12}
		Method     &bicycle   &bird   &boat   & bottle   &bus    & car  &cat
		& chair  &cow  & dog   &mAP \\ \hline
		
		YOLOv3 \cite{redmon2018yolov3}  &$70.25$  &$79.29$  &$53.61$
               &$70.73$   &$93.23$  &$76.48$ &$88.32$
               &$66.51$  &$88.78$ & $84.85$  & $79.08$\\		
		SSD \cite{2016SSD}	&$80.24$  &$72.32$ &$66.30$   &$47.61$ &$83.02$ &$84.21$ &$86.13$ &$54.71$ &$78.36$
		&$84.51$    &$74.63$\\
		Faster R-CNN \cite{girshick2015fast}  &$-$	 &$-$  &$-$ &$-$ &$-$ &$-$ &$-$ &$-$ &$-$ &$-$
		  & $73.24$\\
		YOLOv4 \cite{bochkovskiy2020yolov4}  &$80.80$  &$90.07$   &$75.88$    &$76.71$  &$\bm{95.84}$   &$83.58$  & $92.80$ & $72.38$   & $\bm{95.69}$    & $90.25$     &$85.28$    \\		
		YOLOv5-m  & $82.84$   & $\bm{90.23}$   &$82.27$  &$\bm{82.69}$  &$94.16$  &$93.16$
		&$93.82$   &$74.59$   &$93.38$  &$92.18$   &$87.03$\\
		YOLOv3 ours &$\bm{88.62}$   &$90.02$   &$\bm{82.83}$  &$82.62$  &$95.63$
		&$\bm{94.56}$ &$\bm{93.89}$ &$\bm{75.37}$ &$94.54$ &$\bm{92.63}$   &$\bm{88.76}$  \\			
		\hline
	\end{tabular}
\end{table*}

\section{Ablation Experiments}

In order to verify the influence of different modules in the network on the detection results of small objects, we conducted ablation experiments on the entire network, and the results are shown in Table \ref{tab:abl}.
In this experiment, the interaction between different modules is not considered, and only the influence of a single module on the experimental results is considered.



A $\to$ B First of all, we added a feature enhancement module DCM to the backbone extraction network and the neck. This module effectively improves the context information of small objects in the feature map by fusing feature maps under different receptive fields. As can be seen from the Table \ref{tab:abl}, the detection of small objects in the entire network has increased 5\% after adding the feature enhancement module.

B $\to$ D Next, in order to further enhance the feature extraction ability of the neck in the network, the algorithm proposed in this paper uses a multi-level feature map fusion and attention weighting module to enhance the spatial information of small objects, thereby improving the detection accuracy of small objects. . As can be seen from the results, the module has improved the detection of small objects by 7.8\%.


D $\to$ E In order to further improve the detection accuracy of the network, we use a decoupling head in the network to separately process the regression and classification tasks in the object detection task, thereby improving the detection of small objects by 2.4\%.


E $\to$ F Post processing is also where we can improve the performance, we use the Soft-NMS to replace traditional NMS, and the mAP improved by 0.2\%.
For the detection of small objects, it has improved by 1.3\%.
When the traditional NMS algorithm performs detection box selection, the detection box for some occluded small objects will be removed.
This will cause some small objects  to fail to be detected.
However, the Soft-NMS improves the accuracy of small object detection by reducing the score of occluded small object detection boxes.

\begin{table*}[t]
	\caption{Roadmap of improved YOLOv3 on the COCO 2017 dataset.}
	\label{tab:abl}\centering
	\begin{tabular}{ccccccccc}  \hline
		\cline{1-4}
		&Methods     &AP$_{50:95}$     &AP$_{50}$   &AP$_{75}$   &AP$_{S}$  &AP$_{M}$   &AP$_{L}$  &parameter       \\ \hline
	A	& YOLOv3 baseline   &$38.5$  &$69.3$        &$39.7$         &$16.3$       &$37.9$     &$53.2$  &235M    \\
	B	&+dilated convolution &$41.7$   &$69.2$   &$44.4$  &$21.3$ &$39.1$ &$56.5$  &241M \\
	C	&+attention &$42.3$  &$70.5$  &$45.6$  &$22.5$ &$41.3$  &$55.3$ &245M \\
	D	&+muti-level fusion 	&$43.8$  &$71.1$ &$44.1$   &$29.1$ &$42.5$  &$56.7$ &254M \\
	E	&+decouple head &$46.9$   &$75.4$  &$47.9$   &$31.5$  &$45.3$   &$58.0$  &257M   \\
	F	&+soft-nms  &$ 47.1$    &$76.3$   &$48.5$  &$32.8$   &$46.2$
		&$59.2$  &257M \\
		\hline
	\end{tabular}
\end{table*}

From the above ablation experiments, it can be seen that the use of the dilated convolution module can significantly improve the detection accuracy of large objects,
because the dilated convolution significantly improves the receptive field of the large object detection feature map and get more global information. However, the detection accuracy of small objects is significantly improved by using multi-level feature fusion and attention module. Since this module fuses  different weighted the low-level semantic information and then fuses it with the high-level semantic information obtained by upsampling, so that the entire small object detection head contains more feature information of small objects.

\section{Conclusion}
\label{sec:conclusion}
In this paper, a small object detection model based on YOLOv3 is proposed, which solves the challenge of applying traditional object detection algorithms to small object detection. The main contributions are as follows:

\begin{enumerate}
	\item The DCM module is added to the YOLOv3 network to enrich the information of the feature map and improve the detection accuracy of objects.

	\item  The combination of CABAM and multi-scale fusion module can effectively capture the feature information of small objects and background, suppress non-important information, and effectively improve the detection accuracy of small objects when the model increases a small computational cost.

	\item  The strategy of using Soft-NMS combined with CIoU is beneficial to improve the positioning accuracy of small objects and the detection accuracy of occluded objects.
	
\end{enumerate}

As can be seen from the experimental results of two public datasets,
the method proposed has improved the detection accuracy of small objects by 16.2\% on the original YOLOv3 object detection algorithm. Compared with other algorithms, the algorithm in this paper has improved at least 5\%, without affecting the algorithm for other large and medium object detection. In future work, we will shrink the parameters of the model, so as to realize the real-time detection of the algorithm.

\textbf{\section*{Acknowledgment}}
This work was supported in part by the Gansu Provincial Department of Education University Teachers Innovation Fund Project (No.2023B-056),the Introduction of Talent Research Project of Northwest Minzu University (No. xbmuyjrc201904), and the Fundamental Research Funds for the Central Universities of Northwest Minzu University (No.31920220019, 31920220130), the Leading Talent of National Ethnic Affairs Commission (NEAC), the Young Talent of NEAC, and the Innovative Research Team of NEAC (2018) 98.


\bibliographystyle{spbasic_unsort}
\bibliography{XZ_ref2}
\end{document}